\documentclass[letterpaper, 10 pt, journal, twoside]{IEEEtran} 
\IEEEoverridecommandlockouts

\usepackage{graphicx}
\usepackage{setspace}          
\usepackage{url}
\usepackage{amsmath,bm}
\usepackage{upgreek}
\usepackage{algorithmic}
\usepackage{amssymb} 
\usepackage{mathtools}
\usepackage{xcolor} 
\usepackage{siunitx}
\usepackage{eucal}
\usepackage{systeme}
\usepackage[noadjust]{cite}
\usepackage{subfig}
\usepackage{array}
\usepackage{verbatim}
\usepackage[hidelinks]{hyperref}
\usepackage{caption}
\usepackage[printonlyused]{acronym} 
\acrodef{ROS}{Robot Operating System}

\begin{document}

\title{Load-bearing Assessment for Safe Locomotion of Quadruped Robots on Collapsing Terrain}

\author{Vivian S. Medeiros, Giovanni B. Dessy, Thiago Boaventura, Marcelo Becker, Claudio Semini, \\and Victor Barasuol
\thanks{Manuscript received: June 5, 2025; Revised: August 14, 2025; Accepted: October 9, 2025.}
\thanks{This paper was recommended for publication by Editor Abderrahmane Kheddar upon evaluation of the Associate Editor and Reviewers' comments. This work was supported by the São Paulo Research Foundation (FAPESP) grant 2023/00249-0. (\textit{Corresponding author: Vivian S. Medeiros})}
\thanks{Vivian S. Medeiros is with the Dynamic Legged Systems Lab, Istituto Italiano di Tecnologia (IIT), Genova, Italy, and also with the Department of Mechanical Engineering, São Carlos School of Engineering (EESC), University of São Paulo (USP), Brazil (e-mail: viviansuzano@usp.br).} 
\thanks{Giovanni B. Dessy, Claudio Semini, and Victor Barasuol are with the Dynamic Legged Systems Lab, Istituto Italiano di Tecnologia (IIT), Genova, Italy (e-mail: giovanni.dessy@iit.it; claudio.semini@iit.it; victor.barasuol@iit.it).}
\thanks{Thiago Boaventura, and Marcelo Becker are with the Department of Mechanical Engineering, São Carlos School of Engineering (EESC), University of São Paulo (USP), Brazil (e-mail: tboaventura@usp.br; becker@sc.usp.br).}
\thanks{Digital Object Identifier (DOI): see top of this page.}
}

\markboth{IEEE Robotics and Automation Letters. Preprint Version. Accepted October, 2025}
{Medeiros \MakeLowercase{\textit{et al.}}: Load-bearing Assessment for Safe Locomotion of Quadruped Robots on Collapsing Terrain} 

\maketitle

\begin{abstract}
Collapsing terrains, often present in search and rescue missions or planetary exploration, pose significant challenges for quadruped robots. This paper introduces a robust locomotion framework for safe navigation over unstable surfaces by integrating terrain probing, load-bearing analysis, motion planning, and control strategies. Unlike traditional methods that rely on specialized sensors or external terrain mapping alone, our approach leverages joint measurements to assess terrain stability without hardware modifications. A Model Predictive Control (MPC) system optimizes robot motion, balancing stability and probing constraints, while a state machine coordinates terrain probing actions, enabling the robot to detect collapsible regions and dynamically adjust its footholds. Experimental results on custom-made collapsing platforms and rocky terrains demonstrate the framework's ability to traverse collapsing terrain while maintaining stability and prioritizing safety. 
\end{abstract}


\section{Introduction}
\IEEEPARstart{I}{n} recent years, the deployment of quadruped robots for real-world applications has become a reality, and several commercial robots are already available on the market. 
Significant research efforts in this field are now directed toward making the legged locomotion system more robust to different challenging scenarios by leveraging perceptive and tactile information about the environment.

Navigating challenging terrains is critical in planetary exploration, remote inspections, or construction sites, where unstable or collapsing surfaces are often encountered and pose significant risks~\cite{dettmann22astra,kolvenbach2021}.
To mitigate these risks, a safety assessment of the terrain can be performed by analyzing the terrain map generated by a perception system to identify and avoid potentially unsafe paths~\cite{fankhauser2018,vfa2019,fahmi2022}. Machine learning techniques often segment or classify terrain types from camera images to detect unstable regions~\cite{ginting2023,zhang2024traversabilityawareleggednavigationlearning}. However, using exteroceptive sensing alone for traversability analysis, especially cameras, can be degraded due to poor lighting, reflective surfaces, and sensor occlusion~\cite{focchi2020}. Moreover, because exteroceptive sensors only provide surface-level observations, terrain stability can only be estimated indirectly, often relying on visual appearance rather than physical interaction. 

\begin{figure}[t]
    \centering
    \includegraphics[width=\linewidth]{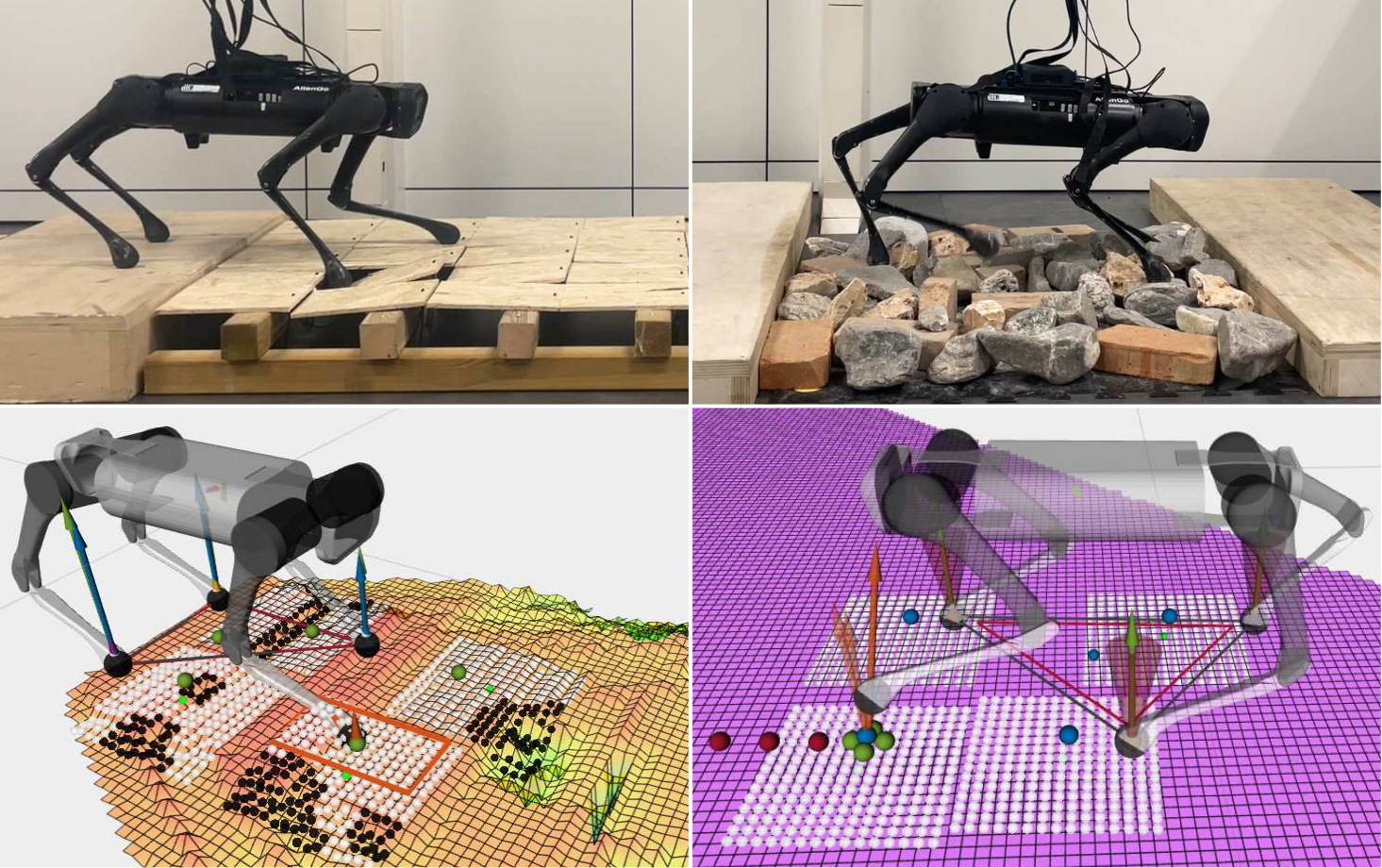}
    \caption{On the top left, the Aliengo robot traverses a customized collapsing terrain with moving planks; on the top right, a terrain with collapsing rocks. On the bottom left, the selection of probing positions is aided by the terrain map (orange square); on the bottom right, the contact force envelope optimized for the probing leg.}
    \label{fig:first}
\end{figure}

For this reason, several previous works have included proprioceptive information for traversability estimation to increase safety in legged locomotion. These approaches combine internal sensing with visual or depth data to enhance terrain understanding, using techniques such as supervised learning~\cite{elnoor2024}, terrain mapping under vegetation~\cite{homberger2019}, multimodal sensing frameworks integrating semantic labels and physical property estimation~\cite{gotafeel}, and reinforcement learning-based control~\cite{anymalLearning}.

While these approaches enhance safety by incorporating physical interaction with the environment, they often focus on surface geometry and assume visually flat terrain is structurally sound. However, even a terrain that would seem safe for foot placement may not support a contact force without collapsing. An example scenario is extra-planetary terrain, often composed of loose soil and sliding rocks. 
In general, terrain probing provides an effective way to analyze terrain stability~\cite{unsupervisedidentification}, which consists of applying a certain force to the terrain before stepping on it to ensure it can support the required forces for locomotion. However, current terrain-probing approaches often rely on either the use of a robotic arm for probing~\cite{gotafeel,TraversabilityAnalysisVisionProbingArm2022} or the installation of special foot sensors, such as force/torque sensors and IMUs, which can be expensive and require changes in the design of the robot's leg~\cite {feetSensorFragileGround,unsupervisedidentification,kolvenbach2019}.
Previous works utilized, e.g., haptic exploration and unsupervised learning for foothold robustness \cite{unsupervisedidentification}; classified granular terrain through impact-induced oscillations measured by foot-mounted force/torque and IMU sensors \cite{kolvenbach2019}; and combined probing with sensor-equipped arms and vision data to create traversability maps \cite{TraversabilityAnalysisVisionProbingArm2022}.

A few approaches have studied legged locomotion in collapsing terrains without using specific force sensors. Haddeler et al.~\cite{haddeler2022} proposed a real-time digital double framework that uses discrepancies between a legged robot and its synchronized digital twin to predict terrain collapsibility. However, the approach only aims to estimate collapse, and no reactive action has been implemented or tested to prevent falls. Conversely, Tennakoon et al.~\cite{probebeforestep} integrate terrain probing into the walking cycle of a hexapod robot, allowing it to test footholds before fully committing to a step. 
The approach was validated on terrains with styrofoam and hidden gaps.

\subsection{Contributions}
Overall, most previous studies in traversability for legged locomotion either do not consider the load-bearing capabilities of the terrain~\cite{elnoor2024,homberger2019,gotafeel,anymalLearning}, or assess terrain collapsibility by using specially designed foot sensors~\cite{feetSensorFragileGround,unsupervisedidentification,kolvenbach2019} or robotic arms~\cite{TraversabilityAnalysisVisionProbingArm2022,gotafeel}.
This paper proposes a legged locomotion system robust to collapsing terrains, integrating perception, planning, and control mechanisms to ensure safe navigation.
Compared to~\cite{probebeforestep}, our approach can traverse more challenging terrains and allows the use of an elevation map of the terrain for foothold planning. The proposed framework also shows improved speed and stability thanks to the aid of MPC. Furthermore, one of the challenges of terrain probing with quadruped robots is the reduced support polygon for base stabilization during probing, which is a less concerning issue for hexapod robots due to the higher number of legs.

To summarize, the main contributions of this paper are:
\begin{itemize}
    \item A motion planning strategy that assesses the terrain load-bearing capabilities by computing the GRF envelope for each leg and probing the terrain accordingly to find safe foothold positions on collapsible terrain. Unlike prior approaches that rely on fixed probing forces, robotic arms, or foot-mounted sensors, the proposed method derives and applies the required probing forces directly from trajectory optimization and joint torque/position measurements, without requiring specialized hardware.
    \item An MPC formulation designed to track the desired motion while balancing support polygon and probing constraints, prioritizing stability;
    \item Experimental validation in rocky terrain and seamlessly flat collapsing terrain with and without terrain mapping. 
\end{itemize}

\section{Locomotion Framework for Load-bearing Assessment in Collapsing Terrain}

\begin{figure}[b]
	\centering
	\includegraphics[width=\linewidth]{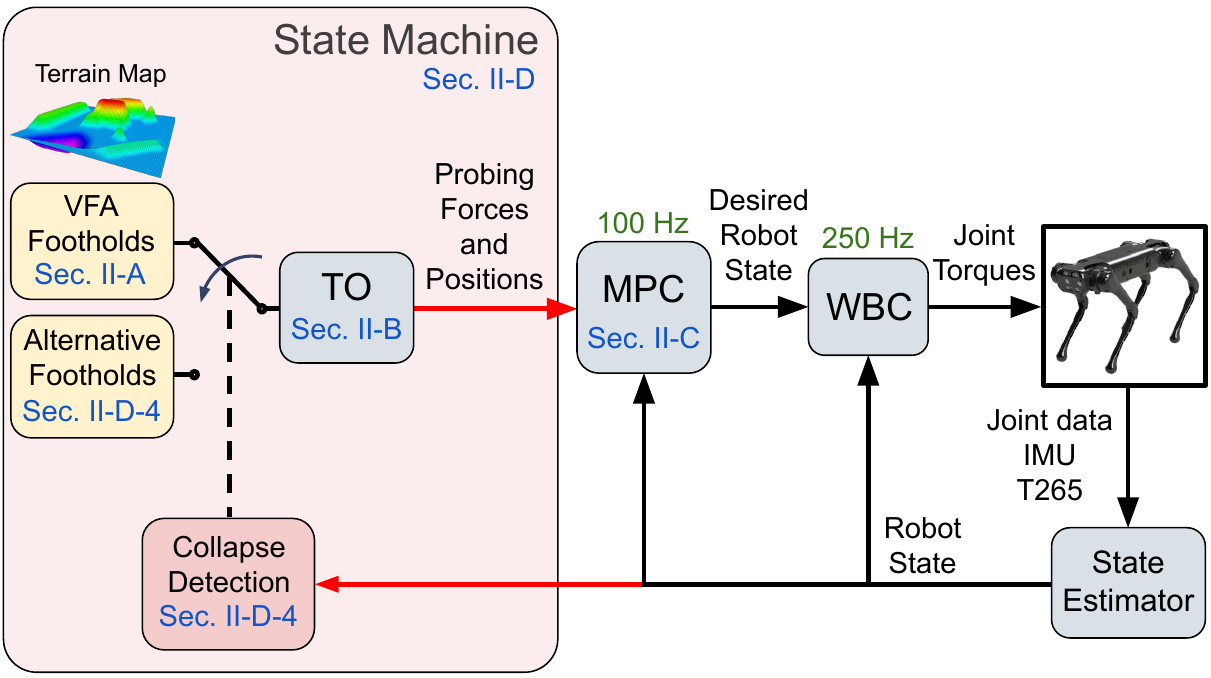}
	\caption{A simplified overview of the proposed framework.}
	\label{fig: overview}
\end{figure}

Figure~\ref{fig: overview} provides an overview of the proposed locomotion framework. Inspired by the strategy humans use when navigating potentially unsafe terrain, the robot probes the terrain before stepping on it to ensure it can support the load of its movement. During this process, the robot applies force to planned foothold positions while stabilizing its base using the remaining three legs.

The magnitude of the probing force at each foothold is computed after optimizing the robot's motion over a full stride and determining the maximum GRF envelope for the probing leg. By avoiding the use of a fixed probing force, this adaptive approach prevents the use of under- or overestimated contact forces, thereby improving energy consumption. If a terrain collapse is detected during contact, the robot quickly removes the foot from the unsafe location, selects an alternative foothold, and repeats the process until a stable position is identified, allowing safe progression.

The selection of probing footholds, collapse detection, and coordination of probing actions are managed by a state machine. Motion planning is handled by an MPC that computes the robot trajectory while enforcing probing-specific constraints such as support polygon and force limits. The use of the MPC allows the system to handle conflicting objectives (stability vs. terrain probing) by weighting them appropriately in the cost function, in addition to improving robustness against model uncertainties. 

The MPC trajectories are tracked by a whole-body controller (WBC) based on virtual Cartesian impedance~\cite{fahmi2019}. The WBC computes the joint actuation torques required to track a desired wrench at the robot center of mass (CoM), derived from the desired trajectories. Additionally, the WBC compensates for the external disturbances estimated by a momentum observer~\cite{focchi2020}.

\subsection{Terrain Mapping and Visual Foothold Adaptation}
\label{sec: vfa}

To improve efficiency, a robot-centric 2.5D elevation map of the terrain~\cite{fankhauser2018} can be used to guide the selection of probing points and eliminate visible unsafe regions. 
Given a set of target foothold positions, a vision-based algorithm analyzes the surrounding grid cells on the map, identifying the closest safe region. The method used for the terrain classification is the Visual Foothold Adaptation (VFA)~\cite{vfa2019}, which classifies the safety of a grid cell based on several heuristic criteria, including terrain roughness, leg collision, and map uncertainty.

\begin{figure}[t]
	\centering
	\includegraphics[width=\linewidth]{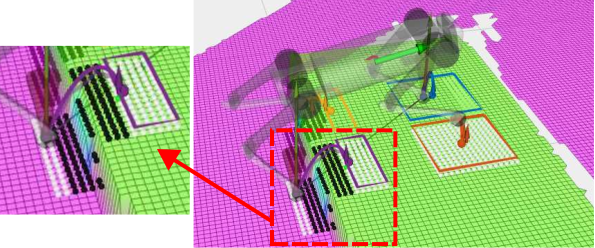}
	\caption{Safe regions for foot placement computed by the VFA algorithm for a scenario with a step. The white spheres indicate the regions of the \textit{foothold heightmap} safe for foot placement, and the black spheres indicate the unsafe regions. They are enclosed by 2D polygons (orange, blue, yellow, and purple), which are used as constraints in the TO.}
	\label{fig: vfa}
\end{figure}

Figure~\ref{fig: vfa} depicts the output of the VFA, which is called the \textit{foothold heightmap}, a 15$\times$15 square and discrete representation of the terrain where each pixel describes the height of the terrain. 
Inside the \textit{foothold heightmap}, the safe foothold positions are indicated with a white sphere, while the unsafe footholds are marked with a black sphere. The VFA also returns the optimal position for foot placement, which is the safe point closest to the desired touch-down position of the leg. 
From the VFA output, it is possible to get the boundaries of the safe region around the target foothold and compute the corresponding 2D convex polygon, which is added as a constraint for foothold placement in the TO.

\subsection{Trajectory Optimization}

A key component of the proposed approach is the optimization of the robot's trajectory for a full stride before each probing step to determine the GRF envelope at the next foothold. The optimization accounts for the movement of the probing leg and the subsequent three legs, all of which affect the forces required at the probing location.

The optimization problem is formulated as:
\begin{subequations}
    \begin{align}
		& \underset{\mathbf{u}(t)}{\text{minimize}}
		&& \int_{t_i}^{t_f} L(\mathbf{x}(t),\mathbf{u}(t),t) \,\text{d}t, &&
		\label{eq:op_cost}\\
		& \text{subject to} && \mathbf{x}(t_i) = \mathbf{x}_i, &&
		\label{eq:initial}\\[-0.5em]
		& && \dot{\mathbf{x}} = \bm{f}(\mathbf{x},\mathbf{u},t), &&
		\label{eq:op_dynamics}\\[-0.5em]
		& && \mathbf{g}(\mathbf{x},\mathbf{u},t) = \mathbf{0}, &&
		\label{eq:op_eq}\\[-0.5em]
		& && \mathbf{h}(\mathbf{x},\mathbf{u},t) \geq \mathbf{0}, && \label{eq:op_ineq}
    \end{align}
    \label{eq:op}
\end{subequations}
\noindent where $\mathbf{x}(t)$ is the robot state, $\mathbf{u}(t)$ is the control input, $\mathbf{x}_i$ is the initial robot state at time $t_i$, $t_f$ is the final time, and $L(\mathbf{x}(t),\mathbf{u}(t),t)$ is the cost function.
The goal is to find the control action that minimizes this cost subject to the initial condition \eqref{eq:initial}, system dynamics \eqref{eq:op_dynamics}, and equality \eqref{eq:op_eq} and inequality \eqref{eq:op_ineq} constraints. 

\textit{1) Inputs}:
the TO receives as input: the robot's current state $\mathbf{x}_i$; the target position and orientation for the robot's base; the reference touch-down positions and the convex safe regions around them, both previously computed by the VFA; the elevation map of the terrain; and the index of the current probing leg to define the contact sequence.

\textit{2) System Dynamics}: the robot dynamics is represented by a Centroidal Dynamics (CD) model~\cite{Orin}.
The state vector is defined as ${\mathbf x = (\mathbf h_{com}, \ \mathbf q_b, \ \mathbf q_j) \in \mathbb{R}^{12 + n_j}}$ and the input vector is ${\mathbf u = (\mathbf f_{c_1}, \ ..., \ \mathbf f_{c_{n_c}}, \ \dot{\mathbf{q}}_j) \in \mathbb{R}^{3 n_c + n_j}}$, where $\mathbf h_{com}$ is the centroidal momentum, $\mathbf q_b$ is base pose (position and orientation), $\mathbf q_j$ are the joint positions, $\mathbf f_{c_{i}}$ are the contact forces and $\dot{\mathbf{q}}_j$ are the joint velocities. The base position and the contact forces are expressed in the world frame, and the orientation of the base is expressed using ZYX-Euler Angles. 

\textit{3) Reference generation}:
the reference trajectory for the robot's base is computed by a linear interpolation between the initial and final positions. Swing reference trajectories are computed between the current and reference touch-down positions as a combination of two quintic splines based on the parameters for the swing motion, such as lift-off velocity, touch-down velocity, and step height. The gait pattern for the locomotion is the \emph{crawl}, with one leg moving at a time, and the total duration of the stride is 4.0~s. Inverse kinematics compute joint position and velocity references corresponding to the desired base and feet trajectories. The reference centroidal momentum is set to zero, and the reference for contact forces is such that it distributes the robot's weight equally to all legs in contact. 

\textit{4) Cost function}:
it is defined by the quadratic cost related to tracking the reference states and inputs for the trajectory:
\begin{equation}
\begin{aligned}
        L(\mathbf{x}, \mathbf{u}, t) = &\frac{1}{2} \left\Vert \mathbf{x}-\mathbf{x}_d  \right\Vert^2_{\mathbf{Q}}  + \frac{1}{2} \left\Vert \mathbf{u}-\mathbf{u}_d \right\Vert^2_{\mathbf{R}} +\\
        &\frac{1}{2} \left\Vert \mathbf{p}_i-\mathbf{p}_{i,ref}  \right\Vert^2_{\mathbf{W}_p} + \frac{1}{2} \left\Vert \mathbf{v}_i-\mathbf{v}_{i,ref}  \right\Vert^2_{\mathbf{W}_v},
	\label{eq:mpc_cost}
\end{aligned} 
\end{equation}
where $\mathbf{R}$, $\mathbf{Q}$, $\mathbf{W}_p$ and $\mathbf{W}_v$ are positive definite weighting matrices.
$\mathbf{x}_d$ is the reference state, $\mathbf{u}_d$ is the reference for the input vector, and $\mathbf{p}_{i}$ and $\mathbf{v}_{i}$ are the positions and velocities for the feet, respectively.

\textit{5) Equality and Inequality Constraints}: for stance legs, the end-effector velocity in the world frame is constrained to zero and a friction cone constraint, $\mathcal{F}$, is enforced defined by the terrain surface normal $\widehat{\mathbf{n}}$ and the friction coefficient, assumed to be ${\mu_c = 0.5}$. For swing legs, zero contact force is enforced, and a reference velocity trajectory $\mathbf{v}_{i,ref}$ is constrained in the normal direction of the terrain, allowing foot placement optimization in the tangential direction.
\begin{equation}
	\left\{ 
	\begin{array}{lll}
		\mathbf{v}_{i} = \mathbf{0},
		\quad  &\mathbf{f}_{c_i} \in \mathcal{F}(\widehat{\mathbf{n}}, \mu_c),
		\quad &\text{if $i \in\mathcal{C}$}, \\
		\mathbf{v}_{i} \cdot \widehat{\mathbf{n}}  = \mathbf{v}_{i,ref}, \quad  &\mathbf{f}_{c_i} = \mathbf{0}, \quad  &\text{if $i \notin \mathcal{C}$},
	\end{array}
	\right.
\end{equation}
where $\mathbf{v}_{i}$ is the end-effector velocity in the world frame, and $\mathcal{C}$ is the set of all legs in contact at a given time.

The foot placement constraint is formulated as a set of linear inequality constraints with respect to the touch-down foot positions $\mathbf{p}_{i}$:
\begin{equation}
	\mathbf{A}_i \cdot \mathbf{p}_{i} + \mathbf{b}_i \geq \mathbf{0},
	\label{eq:perc:foothold_position_constraint}
\end{equation}
where $\mathbf{A}_i \in \mathbb{R}^{4\times3}$, and $\mathbf{b}_i \in \mathbb{R}^{4}$ define $4$ half-space constraints in 3D. Each half-space is defined by an edge of the 2D polygon from the VFA and the surface normal of the corresponding convex region. Other inequality constraints are joint limits (position and velocity) and contact force limits. The inequality constraints are handled as relaxed log-barrier functions in the cost function in the same way as proposed in~\cite{grandia2022}.

In blind locomotion, the VFA is completely disabled. In this case, no convex safe regions are defined around the target touch-down positions, and the trajectory optimization (TO) outputs probing footholds that coincide with the initial targets. The same applies when VFA is enabled but no map is available (e.g., missing sensor data), where the "safe region" defaults to the entire plane.

\begin{figure}[b]
	\centering
	\includegraphics[width=\linewidth]{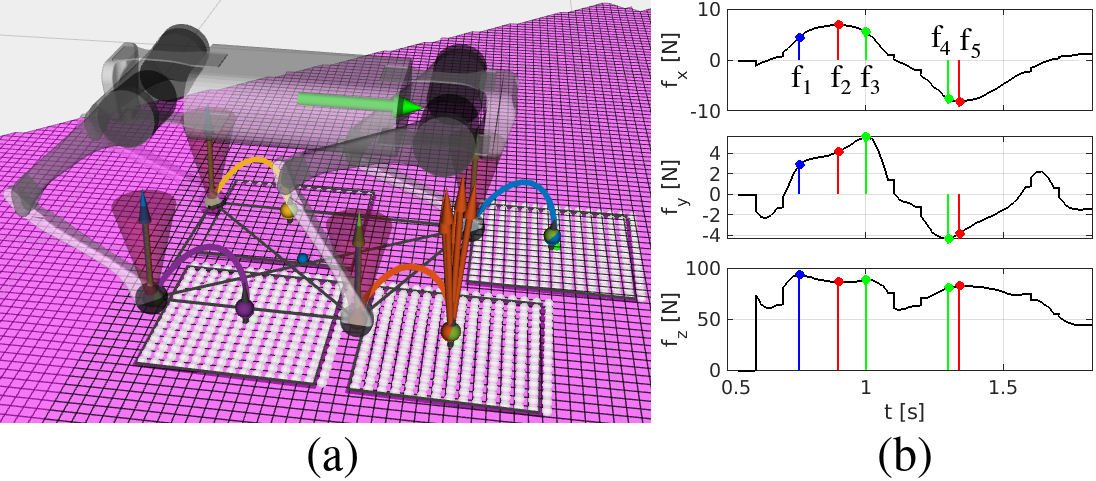}
	\caption{Trajectory optimization: (a) the optimized feet trajectories are represented with colored lines: yellow, blue, purple, and orange. The probing leg is the right-front, and the GRF envelope is indicated by the orange force vectors on the target foot position; (b) the maximum and minimum forces are marked in red, green, and blue, for the x, y, and z-direction, respectively. These forces together compose the GRF envelope. To prevent slippage on the legs, all forces must be contained inside the friction cone, indicated in light red in (a).}
	\label{fig: force_envelope}
\end{figure}

\textit{6) GRF envelope}:
The result from the TO is the whole-body trajectory and control input for the robot's next motion stride. From there, we can extract the VFA-adapted touch-down positions for all legs and the GRF envelope at the next probing position. Figure~\ref{fig: force_envelope}(a) shows the resulting motion for a flat terrain in simulation. The GRF envelope for probing is obtained by computing the maximum and minimum force in each direction ($x$, $y$, and $z$) during the stance phase at the touch-down position of the probing foot, as indicated in Fig.~\ref{fig: force_envelope}(b). 

\subsection{Perceptive MPC} \label{sec:MPC}

Once the TO provides the foot contact positions and forces, an MPC performs real-time motion planning, enabling continuous adjustments based on real-time perception and feedback. At each iteration of the MPC, the optimization problem defined in (\ref{eq:op}) is solved using the most recent state measurement for a 1.0~s horizon. The optimized control policy is applied to the robot until the next MPC update occurs.

The MPC formulation used in this work is based on~\cite{grandia2022}. The main differences are: first, the VFA~\cite{vfa2019} was used for the foothold placement constraints instead of \texttt{elevation\_mapping\_cupy}~\cite{miki2022elevation}; second, the input loopshaping, collision avoidance, and gait adaptations features were not used; and, third, the reference foothold positions are not computed via Raibert heuristic, instead they are defined from the TO. The system dynamics, costs, and constraints remain the same as those for the TO (\ref{eq:op}), but additional constraints are included to handle terrain probing.

The probing procedure involves applying each force of the optimized GRF envelope on the probing position for 100~ms each. During probing, the robot must remain stable to allow quick recovery in case of a collapse. These tasks are accomplished by including two new constraints to the MPC formulation: the support polygon and the probing force constraints. The first one is activated from the moment the probing leg lifts off until the probing of that leg is complete. This helps ensure that the robot is in a statically stable condition until a safe position is achieved. On the other hand, the probing force constraint is only active during the probing phase and imposes the contact forces on a given foot to track a desired force profile. 

\begin{figure}[b]
	\centering
        \includegraphics[width=0.7\linewidth]{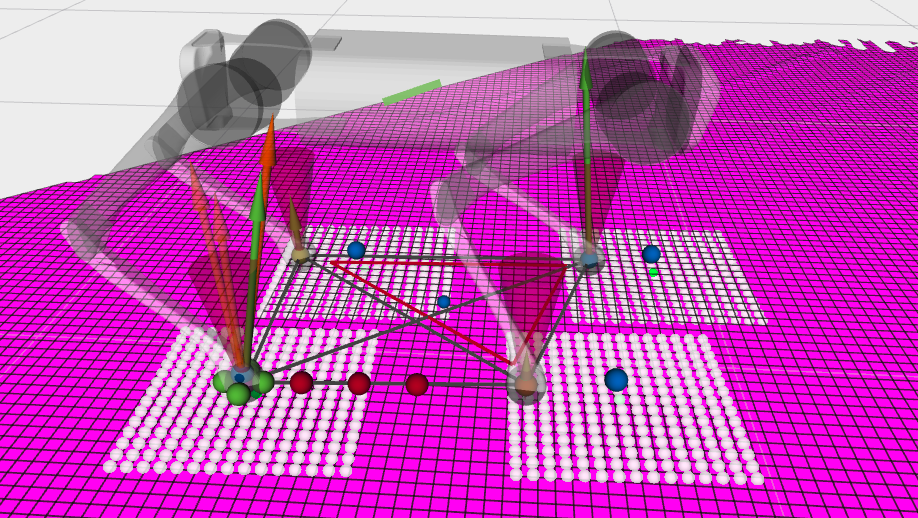}
	\caption{Terrain probing: the blue spheres are the next probing footholds, and the red spheres indicate the alternative probing positions for the current probing leg (right-hind). The orange arrows indicate the GRF envelope computed by the optimization. The limits of the support polygon constraint are indicated with the red polygon. The smaller blue sphere is the projection of the robot's CoM on the terrain in the direction of gravity. The green spheres are the probing-around positions to ensure safety in the region around the probing point. The contact forces are indicated with the green arrows.}
	\label{fig: probing_footholds}
\end{figure}

Figure~\ref{fig: probing_footholds} shows the probing being performed in simulation. Note how the CoM projection (small blue sphere) lies within the limits of the support polygon. Since the probing strategy involves a constant statically stable condition, the support polygon constraint neglects eventual accelerations of the base, and a set of linear inequality constraints gives it:
\begin{equation}
	\mathbf{A}_s (\mathbf{q}_j) \cdot \mathbf{r}_{b}^{x,y} + \mathbf{b}_s - \alpha \geq \mathbf{0},
	\label{eq: support_polygon_constraint}
\end{equation}
where $\mathbf{A}_s$ and $\mathbf{b}_s$ define the linear constraints as functions of the current foothold positions, obtained through forward kinematics from the joint positions $\mathbf{q}_j$; $\mathbf{r}_{b}^{x,y}$ is the 2D CoM position projected in the polygon in the direction of gravity; and $\alpha$ is a safety margin which oﬀsets the support polygon inwards to increase robustness. In case of probing, all the feet are in contact with the terrain, and it should be clear that the concept of support polygon is being extended to include a condition in which all four feet are in the support phase, but the support polygon is defined only by the legs that are not probing the terrain.
The probing force constraint is an equality constraint given by 
\begin{equation}
    \mathbf{f}_i = \mathbf{f}_i^p, \quad  \text{if $i \in \mathcal{C}$},
    \label{fig: probing_constraint}
\end{equation}
where $\mathbf{f}_i^p$ is the desired probing force. This constraint is only activated during the stance phase of the probing foot.

Both the support polygon and the probing force constraints are included in the MPC formulation as soft constraints by adding a penalty to the cost function as a relaxed log-barrier function. As these constraints are opposing, using soft constraints gives flexibility to the formulation for avoiding problems in convergence. Typically, to apply a large force to a foothold position, the base should be moved closer to the foot itself, but the support polygon constraint prevents this from happening. As a result, contact forces are distributed primarily between the two legs forming a diagonal with the probing foot, leaving the other two with low contact forces (Fig.~\ref{fig: probing_footholds}). The parameters of each constraint were adjusted to prioritize the support polygon constraint for safety in the event of collapse. However, this means that the probing force might be slightly lower than required~\cite{dettmann22astra}.
Table~\ref{tab:MPCweights_probing} shows the parameters used for the MPC inequality constraints. The parameters for the soft inequality constraints were tuned as recommended in~\cite{feller2017}. Their work concluded that the barrier function weighting parameter $\mu$ mainly influences the robot's performance, while the relaxation parameter $\delta$ primarily controls the tolerance for constraint violations. The smaller the $\delta$, the closer the constraint is to a hard constraint.
\begin{table}[!h]
	\newcolumntype{C}{>{\arraybackslash}p{4.5cm}}
	\centering
	\begin{tabular}{c|c|c}
		Constraint & $\mu$ & $\delta$   \\
		\hline
		\hline
		Friction cone constraint & 10.0 & 0.1 \\[-0.2ex]
            Foot placement constraint & 0.1 & 0.005 \\[-0.2ex]
		Joint positions limits & 0.1 & 0.01 \\[-0.2ex]
            Joint velocity limits & 0.1 & 0.1 \\[-0.2ex]
            Contact force limits & 1.0 & 0.5 \\[-0.2ex]
            Support polygon constraint & 100.0 & 0.02 \\[-0.2ex]
            Probing force constraint & 0.7 & 0.7 \\[-0.2ex]
	\end{tabular}
	\caption{Barrier function parameters for the inequality constraints in the MPC with the probing-related constraints.}
	\label{tab:MPCweights_probing}
\end{table}

\subsection{State Machine} \label{sec: state_machine}

\begin{figure}[tbh]
	\centering
    \includegraphics[width=\linewidth]{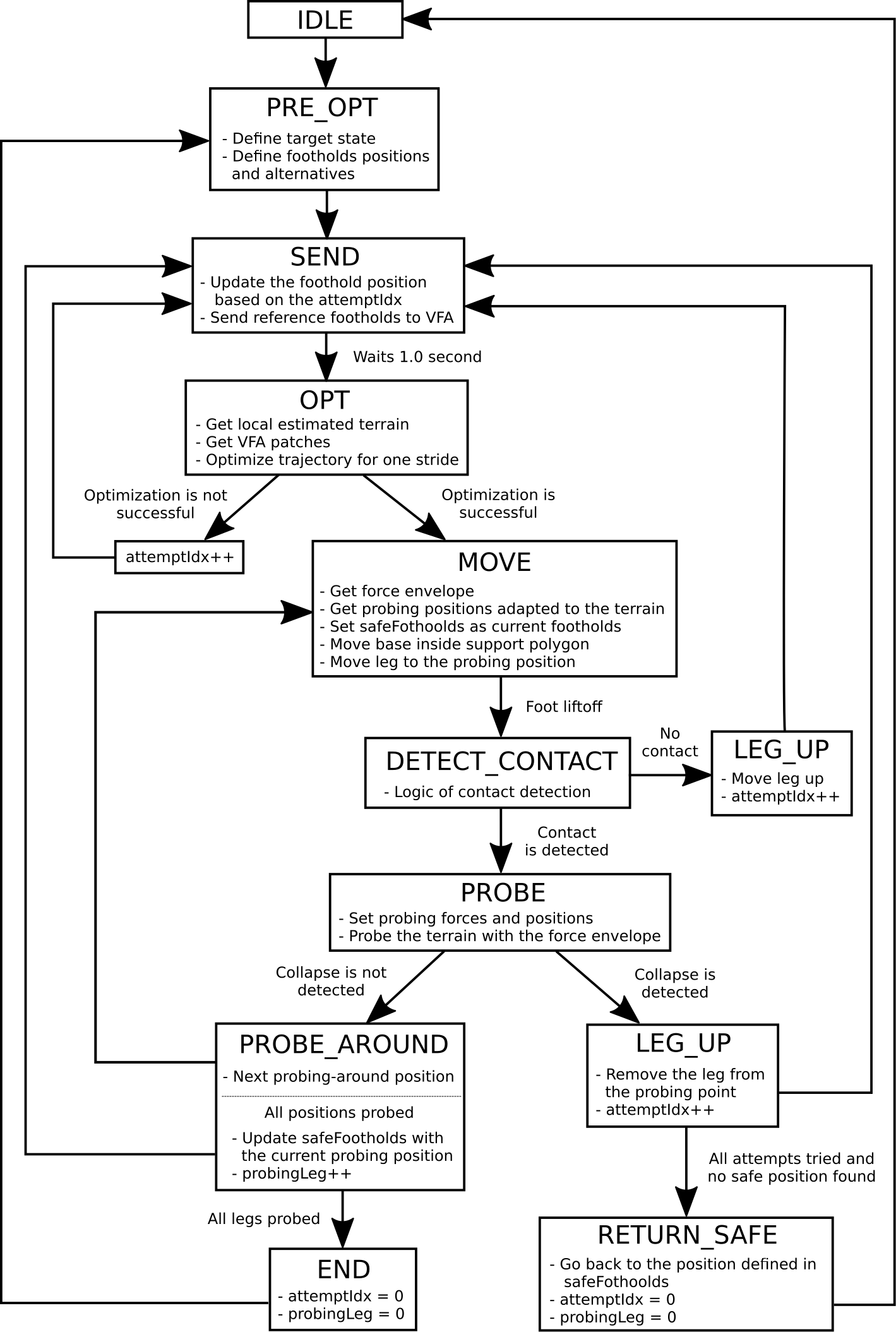}
	\caption{State machine description for the stepping-probing strategy that enables legged locomotion in potentially collapsing terrain.}
	\label{fig: probing_sm}
\end{figure}

A key component of the system is a state machine that manages the probing actions for each leg, detailed in Fig.~\ref{fig: probing_sm}. The main steps of the state machine are described below.

\textbf{1)} PRE\_OPT: The first step is to define the target state for the robot, computed as 10~cm (25\% of leg height) forward in the desired direction of motion, defaulting to the $x$-direction in base frame. Subsequently, the reference foothold positions are defined based on forward kinematics, considering the target base position at a nominal joint configuration. The base height and orientation (pitch and roll) are adapted to the \textit{local surface normal}, which is constantly computed by fitting a plane through the most recent foot contact positions~\cite{gehring2017,focchi2020}. 

\textbf{2)} SEND: Once the target state and the reference footholds are defined, they are given as input to the VFA, which computes the \textit{foothold heightmaps} and the closest safe position for foothold placement, as described in Sec. \ref{sec: vfa}. When used, visual-based terrain mapping only accelerates foothold selection, while probing guarantees load-bearing safety of the terrain.

\textbf{3)} OPT: In this step, a TO that considers the VFA terrain information for foot placement is performed, computing the ideal probing position and GRF envelope required for that point to withstand the robot's motion. 
The output is not only the maximum force magnitude, but also its direction, which increases the chances of collapse detection during probing.

After the optimization, the terrain-adapted probing position and the GRF envelope are provided as input to the MPC, which runs continuously at 100~Hz in parallel to the state machine. Before moving the leg to the probing point, the current foot positions are stored as the latest safe positions on the terrain. Each safe foothold is only updated after a successful probing. This is important to enable the robot to return to a safe condition if all attempts to find a non-collapsible foothold fail. Furthermore, three alternative foothold positions, each spaced 7.5~cm apart\footnote{This parameter was defined based on the size of the robot, specifically the distance between front and hind legs, and might require adjustments for larger or smaller robots.}, are defined in front of the probing position in case of collapse, indicated with the red spheres in Fig.~\ref{fig: probing_footholds}.

\textbf{4)} MOVE and DETECT\_CONTACT: The next step is to move the foot to the respective probing position (MOVE). First, the robot moves its base into the support polygon formed by the other three legs. The leg starts the swing trajectory once the base reaches a safe position. As soon as the foot lifts off from the ground, the algorithm triggers a state that waits for a contact detection (DETECT\_CONTACT). If the contact is supposed to happen, but no contact is detected, the foot continues descending until contact is measured or the maximum leg extension is reached. If, on the other hand, an early contact is detected, the probing process starts in the detected contact position. This logic is particularly useful when performing the load-bearing assessment blindly, without terrain information. 
However, drifts in the state estimation and the elevation mapping make this logic useful even when exteroceptive information is available. Furthermore, the foot position at the moment of contact is essential for collapse detection during probing. 

Considering that the robot is in quasi-static condition during the load-bearing assessment, the force estimate at the contact point can be obtained by the full rigid-body dynamics of the quadruped robot~\cite{camurri2017} as:
\begin{equation} 
	\mathbf {f}_{i} = -\big(\mathbf{J}_{i}^{\rm{T}}(\mathbf {q}_{i})\big)^{-1} \big(\boldsymbol{\tau}_{i} - \mathbf {h}_{i} (\mathbf {q}_{i},\dot{\mathbf {q}}_{i}, \mathbf{g})\big) 
\end{equation}
\noindent where $\mathbf{J}_{i}^{\rm{T}}(\mathbf {q}_{i})$ is the Jacobian transpose from joint to Cartesian space, $\mathbf {h}_{i} (\mathbf {q}_{i},\dot{\mathbf {q}}_{i}, \mathbf{g})$ is the vector of centrifugal, Coriolis and gravity torques for leg $i$, and $\boldsymbol{\tau}_{i}$ are the joint torques for that leg. Contact is detected if the normal component of the contact force is above a predefined threshold.

Moreover, the foot is considered in contact only after three successive positive contact measurements, which helps prevent false contacts from noise data or oscillations that arise naturally from the impact with the terrain.
In case of leg over-extension and no contact detection, the leg is kept in the air until the next alternative probing position is computed. If a contact is detected, the terrain probing is initiated again.

\textbf{5)} PROBE: In this step, the support polygon and probing constraints are enabled in the MPC, as described in Sec.~\ref{sec:MPC}. If all the forces in the envelope are applied to the terrain and no collapse is detected, an additional measure of safety is employed to make sure that not only is the foothold position stable, but also the region around it (PROBE\_AROUND). For that, the robot also probes four points around the original foothold (front-back-left-right) in a 2.5~cm radius, which is determined by the radius of the robot's footpad. The terrain is considered safe only if no collapse is detected at all probing positions. This process is repeated for the next leg until all the legs have probed the terrain and the robot has completed a full stride cycle. This continues until the user stops the probing state machine. One downside of this process is that the robot moves at a much slower velocity to probe all points in the terrain before moving. Ideally, potentially dangerous or collapsible terrain regions could be predetermined so that the robot can apply the stepping-probing strategy only when necessary.

Collapse is detected based on the displacement of the foot during force probing. The foot position is stored after the leg moves to the probing point (MOVE) and contact is detected (DETECT\_CONTACT). If, during probing, the difference between the current foot position and the stored one is higher than 3~cm\footnote{This parameter was defined empirically based on collapsing tests in different terrains, including foam, paper, and moving rocks.}, terrain collapse is assumed and the leg is removed from the probing point and kept elevated until a new probing point has been chosen by the TO (LEG\_UP). For the LEG\_UP motion, a new reference state for the robot is provided to the MPC in which the robot base is positioned inside the support polygon, and the probing leg is elevated at a fixed position below the hip.

In a condition where none of the alternative probing points in the vicinity of the desired foot positions are safe, the leg returns to the previous safe position (RETURN\_SAFE), and the state machine is stopped until the user or a high-level path planning system provides a new direction of motion.

\section{Results}

\subsection{Implementation}

The entire framework was implemented in C++, and ROS was used to communicate between the MPC and the WBC.
For both the TO and the MPC, the optimal control problem is solved using a multiple shooting approach based on Sequential Quadratic Programming (SQP). The toolbox for solving the optimal control problem is OCS2~\cite{OCS2}.
The time horizon for the MPC was set as 1.0~s, with a sampling time of 0.015~s, and the MPC frequency was set to 100~Hz, which was fast enough to recompute the motion plans and react to collapse detections. On average, the computational cost for solving the TO was 100~ms for optimizing the entire motion. The WBC runs at 250~Hz, and the desired contact force computed by the MPC is provided as a bias for the computation of the joint actuation torques.

\subsection{Experimental Results}
\begin{figure}[b]
	\centering
        \includegraphics[width=\linewidth]{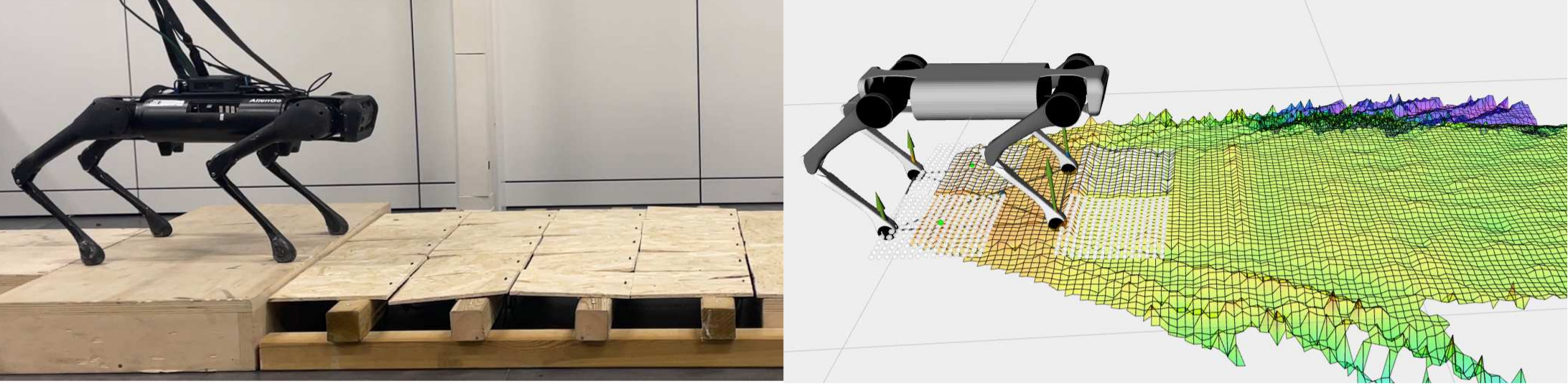}
	\caption{On the left, the custom-made collapsing terrain; on the right, the corresponding elevation map shows a seemingly stable and flat surface.}
	\label{fig: collapsing_terrain_map}
\end{figure}

The proposed approach was validated with simulations and experimental tests in collapsing terrain with and without mapping information\footnote{Video:\url{https://www.youtube.com/watch?v=uTiwrigGOVU}}. Two main scenarios were tested: a custom-made platform with tilting planks and a terrain with unstable and collapsing rocks.
The custom-made terrain is composed of wooden planks attached to a central beam. Each plank is 200$\times$250~mm, and the terrain is composed of 4 rows each, with a total length of 1~m. 
A spring is attached at the end of each plank, so the plank acts as a lever when a force is applied. 
This is an adequate scenario for testing because the VFA would not help in avoiding unsafe parts of the terrain, since the planks are arranged uniformly and close enough to each other that the map hardly detects any gap between them, as shown in Fig.~\ref{fig: collapsing_terrain_map}.

\subsubsection{Terrain with tilting planks}

\begin{figure}[b]
	\centering
    \includegraphics[width=\linewidth]{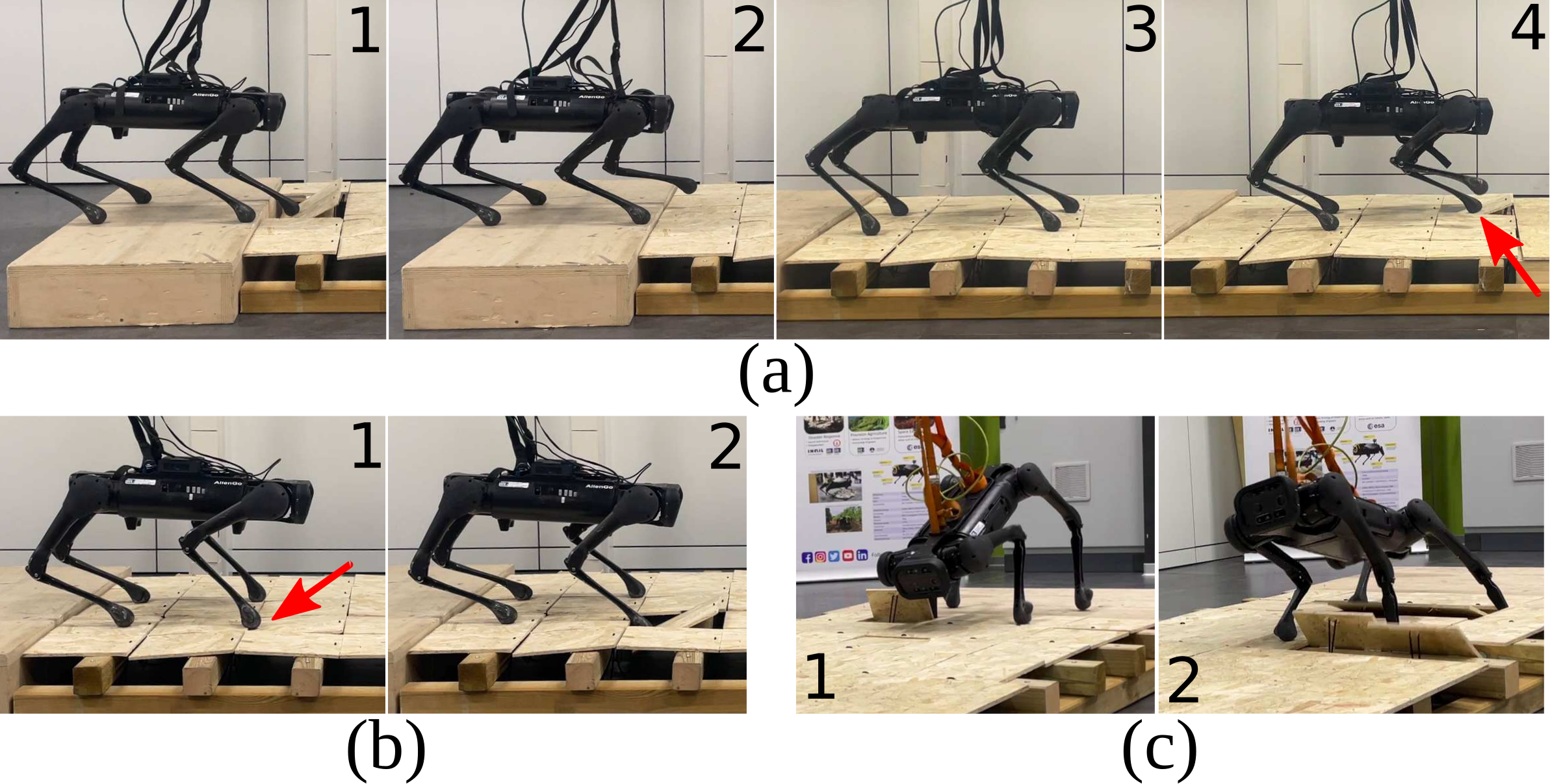}
	\caption{(a) Snapshots of the robot successfully traversing the collapsing terrain with tilting planks; (b) Probing around strategy prevented the robot's foot from being placed in a region too close to an edge; (c) Fail in traversal using the baseline MPC approach.}
	\label{fig: collapsing_terrain_traversal}
\end{figure}

The robot could cross the collapsible terrain both in blind locomotion and with the VFA information from the elevation mapping. The traversal was concluded in 6.5~minutes, on average\footnote{Average computed from 5 complete traversals for the robot starting at different initial positions.}, while maintaining the expected stability along the trajectory. Figure~\ref{fig: collapsing_terrain_traversal}(a) shows snapshots of the traversal.
Snapshots~1 and~2 show the collapse reaction: in 1, the robot starts probing the terrain on the collapsible part of the plank, causing it to move; and in 2, the collapse is detected and the robot immediately removes the foot from the ground while maintaining the base in a stable condition. Snapshots~3 and~4 show that the robot could find stable foothold positions along the terrain to support its locomotion. As expected, most safe positions were encountered on top of the beams that hold the planks, as shown in snapshot~3. However, due to the knowledge of the terrain's collapse limit, the robot can also balance itself and keep stable in the transition area in the vicinity of the beam (snapshot~4).
Figure~\ref{fig: collapsing_terrain_traversal}(b) shows a condition where the probing-around strategy prevented the foot from being placed dangerously close to the edge of the plank, which could have caused the robot to fall. When the robot probed the point forward to the original probing foothold, a collapse was detected, and a new foothold was chosen. Lastly, Figure~\ref{fig: collapsing_terrain_traversal}(c) shows the baseline approach, which relies solely on the MPC without probing constraints, failing to traverse the terrain as the robot's legs became trapped beneath the planks. This comparison demonstrates that an MPC-only controller, even though effective on many rough terrains, fails in collapsing scenarios where dynamic recovery is not possible, whereas the proposed framework succeeds.

\subsubsection{Terrain probing}
\begin{figure}[t]
	\centering
	\includegraphics[width=\linewidth]{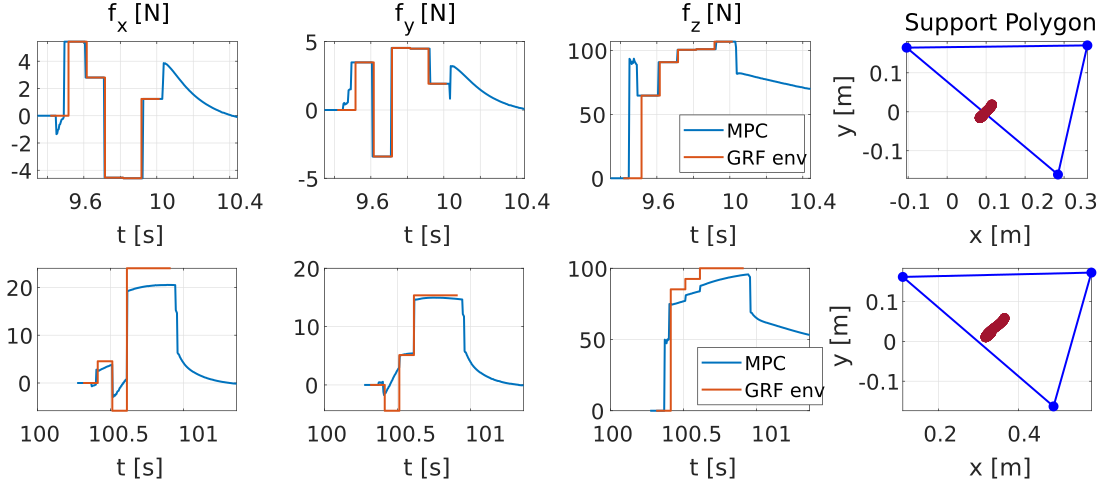}
	\caption{Comparison plots between the optimized GRF envelope, in orange, and the optimized contact force generated by the MPC and provided as input to the WBC (in blue). On the right, the polygon formed by the non-probing legs (in dark blue) and the robot's CoM projection (dark red dots). The first row shows the results without the support polygon constraint during probing, and the second row shows the results with it.}
	\label{fig: force_plots_sim}
\end{figure}

\begin{figure}[t]
	\centering
	\includegraphics[width=\linewidth]{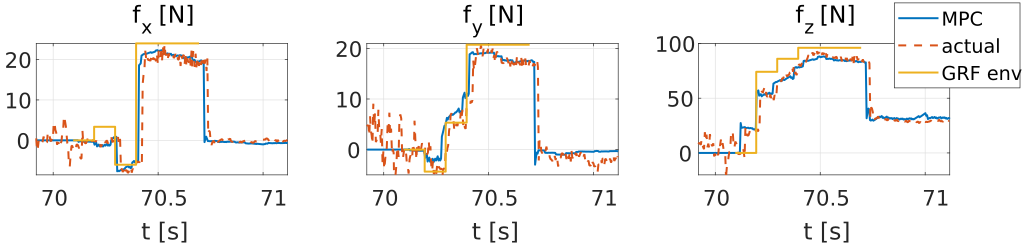}
	\caption{Comparison plots between the optimized GRF envelope, in orange, the optimized contact force by the MPC, in blue, and the actual force obtained from joint torque measurements from the real robot, in dashed red.}
	\label{fig: force_plots_real}
\end{figure}

The core part of the proposed approach is probing the terrain with the actual required GRF envelope. However, a trade-off exists between GRF tracking and the robot's stability. Figure \ref{fig: force_plots_sim} shows a comparison of the contact forces computed by the MPC during leg probing in non-collapsible terrain without the support polygon constraint (top row) and with it (bottom row). Without the support polygon constraint, the MPC can fully track the optimized GRF envelope; however, the robot's CoM projection can move outside the polygon defined by the non-probing legs, which would cause failure to react to terrain collapse. Adding the support polygon constraint ensures safety by maintaining the CoM within a stable region, but it increases the tracking error of the GRF envelope during probing. Nonetheless, the experiments confirm that the applied probing force remains sufficient to detect collapsible terrain and prevent falls. Furthermore, the WBC was able to track the desired contact forces successfully, as shown in Fig. \ref{fig: force_plots_real}. One approach tested to reduce the GRF envelope tracking error while preserving the necessary support polygon constraint was reducing the base pose tracking gains during the probing phase. This would allow higher angular acceleration of the base, enabling the desired contact force to be applied at the foot while keeping the CoM projection within the support polygon. However, the resulting high acceleration references caused excessive motor current and power limits on the real hardware, a limitation of the experimental platform used in this study.

\subsubsection{Terrain with missing planks}
\begin{figure}[t]
	\centering
	\includegraphics[width=\linewidth]{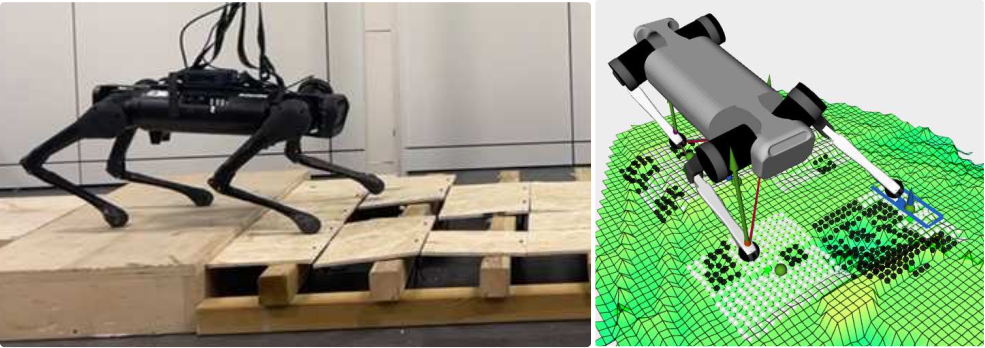}
	\caption{The robot traverses the terrain with some planks removed. On the right, it is visible that the hole in the terrain is marked as unsafe by the VFA, and the foothold selection (blue square) was on the traversable part of the terrain, outside the hole.}
	\label{fig: vfa_terrain_holes}
\end{figure}

Figure~\ref{fig: vfa_terrain_holes} shows a scenario in which the VFA could help with foothold placement before probing by identifying unsafe terrain regions. A few planks were removed from the custom-made collapsible terrain so the VFA could locate them as not traversable and guide the foothold selection. Note how probing is still necessary since the safe region selected by the VFA is still collapsible.

\subsubsection{Terrain with loose rocks}
\begin{figure}[t]
	\centering
	\includegraphics[width=\linewidth]{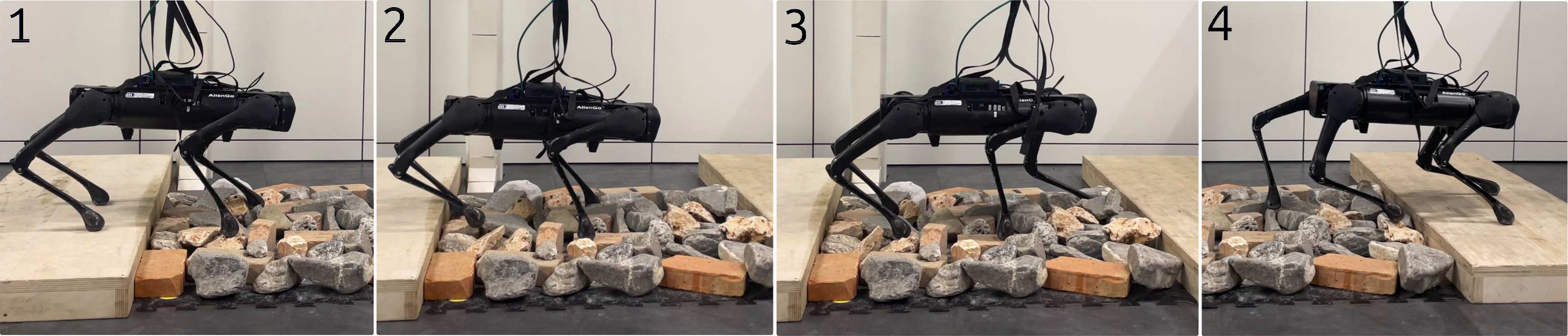}
	\caption{The robot successfully traverses loose rock terrain.}
	\label{fig: rocks}
\end{figure}

One important collapsing scenario encountered in real-world applications is a terrain with loose rocks and bricks. Figure~\ref{fig: rocks} shows the robot traversing such terrain in the most challenging condition, blind locomotion. Since the stones are loose, the robot probes several positions until it finds a safe one and keeps itself balanced on the rocks throughout the path.

\section{Conclusions}
This paper presented a novel framework for safe quadruped robot locomotion on collapsing terrains, addressing the challenges of terrain stability. By integrating trajectory optimization and MPC, the proposed approach assesses the load-bearing capacity of the terrain by computing the GRF envelope for each leg and probing the surface to find safe foothold positions on collapsible terrain without requiring specialized hardware. A state machine coordinates probing actions and collapse recovery, ensuring continuous operation even in unpredictable environments.
Experimental results demonstrated successful traversal of collapsing platforms and rocky terrains, with and without the terrain elevation map. The framework improved stability, reduced the risk of falls, and optimized locomotion performance by incorporating both proprioceptive and perceptive feedback.

Future work will explore scaling the method to higher-speed locomotion and leveraging visual semantics to reduce unnecessary probing. Unlike the current VFA, which relies on geometric heuristics from elevation maps, semantic classification (e.g., rock, sand, vegetation) could help filter out obviously unsafe regions and prioritize where probing is most valuable. Probing, however, remains essential for load-bearing verification in rocky or visually flat terrains, where visual cues alone are insufficient.

\bibliographystyle{IEEEtran}

\end{document}